# Unsupervised Multi-Clustering and Decision-Making Strategies for 4D-STEM Orientation Mapping

**Junhao Cao[1,2], Nicolas Folastre[1,2], Gozde Oney[3], Edgar Rauch[4], Stavros Nicolopoulos[5], Partha Pratim Das[5], Arnaud Demortière[1,2,6*]**

[1]Laboratoire de Réactivité et Chimie des Solides (LRCS), CNRS UMR 7314, Université de Picardie Jules Verne, Hub de l'Energie, Rue Baudelocque, 80039 Amiens Cedex, France.

[2]Réseau sur le Stockage Electrochimique de l'Energie (RS2E), CNRS FR 3459, Hub de l'Energie, Rue Baudelocque, 80039 Amiens Cedex, France.

[3]Institut de Chimie de la Matière Condensée de Bordeaux (ICMCB), Bordeaux, France.

[4]Université Grenoble Alpes, CNRS, Grenoble INP, SIMAP, 38000 Grenoble, France

[5]NanoMegas company, Belgium

[6]ALISTORE-European Research Institute, CNRS FR 3104, Hub de l'Energie, Rue Baudelocque, 80039 Amiens Cedex, France.

Corresponding Author: *arnaud.demortiere@cnrs.fr

## ABSTRACT

This study presents a novel integration of unsupervised learning and decision-making strategies for the advanced analysis of 4D-STEM datasets, with a focus on non-negative matrix factorization (NMF) as the primary clustering method. Our approach introduces a systematic framework to determine the optimal number of components (k) required for robust and interpretable orientation mapping. By leveraging the K-Component Loss method and Image Quality Assessment (IQA) metrics, we effectively balance reconstruction fidelity and model complexity. Additionally, we highlight the critical role of dataset preprocessing in improving clustering stability and accuracy. Furthermore, our spatial weight matrix analysis provides insights into overlapping regions within the dataset by employing threshold-based visualization, facilitating a detailed understanding of cluster interactions. The results demonstrate the potential of combining NMF with advanced IQA metrics and preprocessing techniques for reliable orientation mapping and structural analysis in 4D-STEM datasets, paving the way for future applications in multi-dimensional material characterization.

**INTRODUCTION**

Recent advancements in scientific instruments for material analysis have led to development of devices enable to generate vast amounts of data across multiple modalities with high spatial resolution. These large and complexed datasets often require advanced AI algorithms for efficient processing. For instance, in transmission electron microscopy (TEM) field, the recent integration of advanced techniques such as hybrid-pixel detection, electron beam precession and highly coherent beam has culminated in the emergence of a new class of hyperspectral analysis known as four-dimensional scanning transmission electron microscopy (4D-STEM) [1a]. The 4D-STEM technique involves acquiring a two-dimensional dataset of diffraction patterns over a two-dimensional scanning region, resulting in a four-dimensional dataset [1b]. A single 4D-STEM dataset can contain more than 100 k electron diffraction patterns ($512*512$ $px^2$). Consequently, storing, managing, and efficiently analyzing such data presents significant challenges. The complexity of 4D-STEM data stems from the multidimensional nature of the structural information encoded within each diffraction pattern [24], involving advanced algorithms and significant computational resources for effective analysis. Moreover, preprocessing electron diffraction patterns to maintain data integrity and mitigate various noise sources introduces significant computational overhead and remains a time-consuming task. Consequently, automated data processing leveraging advanced algorithms, particularly those integrating statistical methods and machine learning techniques, is crucial for enhancing efficiency and accuracy in diffraction pattern analysis.

Pattern matching strategies based on pixel-to-pixel cross-correlation coefficients between experimental patterns and simulated patterns, generated from known crystallographic structure data in Crystallographic Information File (CIF) format [2a], have been extensively employed for the analysis of 4D-STEM datasets. This method, which facilitates the extraction of orientation and phase maps, has been implemented in several software packages, including Astar [2b], py4D-STEM [2c], and pyXEM [2b]. The automated crystal orientation mapping (ACOM) procedure determines the orientation of each diffraction pattern, enabling accurate crystallographic analysis of materials. However, electron diffraction patterns are inherently sparse datasets, with fewer than 10% of the pixels containing meaningful signal. Thus, the implementation of data reduction strategies, which convert sparse data into dense representations, can significantly enhance post-processing efficiency for feature extraction, clustering, and reconstruction, as demonstrated in the development of ePattern [42].

Clustering and data reduction strategies are standard techniques for handling large and high-dimensional datasets. Their objective is to enhance data interpretability while preserving the most relevant information from the original dataset [3] [4]. For instance, Principal Component Analysis (PCA) is a widely used unsupervised learning technique for dimensionality reduction, transforming data into a new coordinate system to capture most of the variance in fewer dimensions [4,5,6]. While effective in applications like image processing, noise reduction, and data compression, PCA has limitations, including its inability to capture non-linear data structures and the interpretability challenges posed by negative component values [11,12]. Furthermore, when combined with clustering algorithms, PCA's results can be sensitive to the user-defined number of clusters, potentially affecting analysis robustness [13].

In contrast, Non-negative Matrix Factorization (NMF) [10a] offers several advantages over PCA in the context of unsupervised learning and data dimensionality reduction. Unlike PCA, which allows for both positive and negative components, NMF imposes non-negativity constraints on the factorized matrices. This non-negativity constraint results in a parts-based data representation, making NMF highly effective for interpreting and extracting meaningful features in applications such as image processing, text mining, and spectral data analysis. Furthermore, NMF is better suited for handling non-linear and non-convex data structures because it does not assume data linearity inherent in PCA. The additive nature of NMF components can capture the underlying data patterns more effectively when the data consists of overlapping or additive features. NMF's powerful ability to extract subtle orientation variations has been utilized to enhance the accuracy and reliability of detecting different crystal orientations in 4D-STEM datasets [27].

In traditional clustering methods, the determination of the optimal number of clusters is inherently challenging due to several factors [27b]. The intrinsic complexity of the data can make the natural separations between clusters unclear, especially in the presence of overlapping clusters, noise, or varying density and shape [10b]. The absence of ground truth in many clustering applications requires reliance on data-driven methods to estimate the optimal number of clusters. To tackle these challenges, various methods have been proposed. The elbow method entails plotting the within-cluster sum of squares (WCSS) against the number of clusters to identify a point where adding more clusters yields diminishing returns [10c]. Silhouette analysis assesses cluster compactness and separation, selecting the number of clusters that maximizes the silhouette score [10d]. Incorporating domain knowledge can also guide and validate the clustering process, ensuring alignment with practical expectations [10e]. By integrating these approaches and validating results across multiple criteria, the determination of the optimal number of components in clustering becomes more robust and reliable.

Brute-force or sophisticated methods for determining the optimal number of clusters usually involves running the clustering algorithm multiple times, each with a different number of clusters, and selecting the configuration that yields the most favorable results. These approaches are computationally intensive. To address this issue more effectively, integrating decision-making approaches, such as multi-criteria decision-making techniques, can provide substantial advantages by automating the selection process and enhancing the robustness of the clustering outcomes. Decision-making can be considered as a problem-solving method providing an optimal solution to a specific event [14,15]. After analyzing of a finite set of alternative solutions, the objective is to categorize these alternatives to establish a priority ranking among them. Generally, the conception of decision-making in unsupervised learning [17] is related to extracting significant patterns, features, or underlying information [16], without specific labels, revealing the inherent characteristics or relationships hidden in the raw data [18,19]. In the 4D-STEM data clustering process, decision-making involves several considerations specific to the qualities and attributes of electron diffraction pattern datasets, which encompass both crystal orientation and crystallographic phase information [20].

An additional significant challenge in 4D-STEM mapping is the overlap of patterns from different crystals [20, 20b]. In 4D-STEM, diffraction patterns are generated from probe positions scanning crystals that may be in proximity or/and superimposed configurations [2, 20]. Thus, assigning the correct crystallographic orientation becomes difficult when overlapping occurs [25]. The diffraction patterns in 4D-STEM can demonstrate complicated features and overlapping spots, the ambiguity of which leads to requiring accurate interpreting of the orientation [24]. The complexity of diffraction patterns can cause errors or uncertainties regarding the determining crystal orientations [26, 27]. Efficient algorithms for overlap detection are thus required to specify the precise location of each individual diffraction pattern [25, 25b].

In this study, we develop clustering approach using Non-negative Matrix Factorization (NMF) to analyze four-dimensional scanning transmission electron microscopy (4D-STEM) datasets for orientation mapping. We introduce an efficient method termed "K-Component Loss," which, when combined with Image Quality Assessment (IQA), enables the automatic and effective detection of material characteristics and clustering within large datasets. Our methodology begins with an evaluation phase (Level One) to determine initial NMF parameters. Then, we employ a k-metric derived from IQA to ascertain the optimal number of clusters (k) in a subsequent phase (Level Two). This approach is particularly advantageous for processing overlapping diffraction patterns, as it leverages advanced data analysis techniques to separate overlapping signals, assess the similarity of each component, and accurately extract pertinent features from the dataset. By integrating NMF with IQA, our making-decision method offers a robust framework for the analysis of complex 4D-STEM data, facilitating enhanced material characterization and more precise orientation mapping.

## METHODS

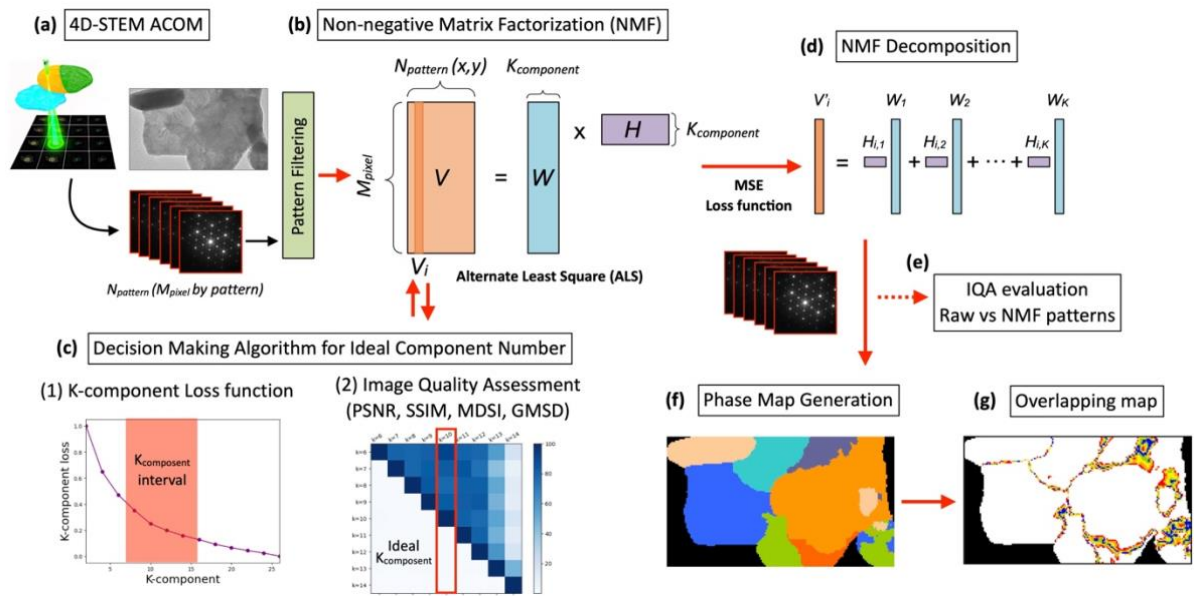

**Figure 1.** Schematic representation of the global workflow for clustering and decision-making strategies in the analysis of the 4D-STEM dataset. **(a)** Overview of the 4D-STEM ACOM acquisition methodology. **(b)** Hypermatrix decomposition of the original dataset into two matrices, WWW and HHH, using the Non-Negative Matrix Factorization (NMF) algorithm. **(c)** Core decision-making framework for determining the optimal number of clusters (components): ($c_1$) Initial step identifying the potential range of cluster numbers, and ($c_2$) Similarity assessment of pattern pairs to maximize differentiation while avoiding overfitting. **(d)** Reconstruction of diffraction patterns based on the NMF-derived results. **(e)** Image Quality Assessment (IQA) comparing raw diffraction patterns to those reconstructed via NMF. **(f)** Visualization of individual clusters (components) within the dataset, aligned with the optimal component count. **(g)** Creation of an overlapping map highlighting the regions of cluster co-occurrence.

### Non-negative matrix factorization algorithm

Non-negative matrix factorization (NMF) [1,2] is a common unsupervised machine learning algorithm that decomposes an original non-negative matrix *V* into two non-negative matrices, *W* and *H*. Popularized by Lee and Seung in 1999 [29], NMF was initially applied in image processing to achieve parts-based representations of face images by combining learned features. Since its introduction, non-negative matrix factorization (NMF) has become a powerful unsupervised learning algorithm, particularly valued for its superior interpretability in uncovering latent features. By decomposing data into non-negative components, NMF facilitates the identification of meaningful patterns, enhancing the understanding of underlying structures in various datasets.

In the latent space, essential features of the original matrix are extracted by selecting components (denoted as k) whose number is significantly less than the rank of the original matrix $V$ (k ≪ min($W, H$)). The matrix $V \approx W * H$ is factorized into two relatively small matrices *(W, H)* compared with V (Original), the dimensionality of these two matrices is $W * k$ and $k * H$, respectively [30]. The linear combination of *W* and *H* generates an approximated matrix $V' = W * H$. *W* matrix can be interpreted as the feature matrix, in which the *k*-column represents the most *k*-relevant feature from the original matrix *V*. *H* can be interpreted as the coefficient matrix, in which the element is the weight associated with the W matrix. Moreover, the aim of obtaining the result of approximate matrix V' is achieved by minimizing a loss function [29].

Lee and Seung introduced an alternating optimization method for NMF [29]. Starting with random non-negative initializations of matrices W and H, the algorithm iteratively, Alternating Least-Square (ALS), minimizes the loss function ||V – WH|| using multiplicative update rules. In each iteration, H is updated while keeping W fixed, followed by updating W with H fixed, ensuring that both matrices remain non-negative throughout the process. This procedure continues until the difference between V and its approximation WH falls below a predefined threshold [30].

**Data preparation**

Non-negative Matrix Factorization necessitates a two-dimensional (M×N) non-negative input matrix. Given that 4D-STEM datasets are inherently four-dimensional, with dimensions (M, N, x, y), in which (M, N) represent probe positions and (x, y) correspond to pixels within each diffraction pattern (512*512 $px^2$), it is imperative to preprocess these datasets appropriately. Typically, the product M×N correlates with the dataset's size [1]. Therefore, converting the 4D-STEM dataset into a two-dimensional matrix V is essential for subsequent matrix computations (details in SI).

To ensure dataset integrity following Non-negative Matrix Factorization (NMF), it is essential to assess information loss between the original and factorized matrices. Incorporating L1 regularization [32], commonly utilized in machine learning to enhance model sparsity, can effectively select pertinent features, particularly in high-dimensional datasets [33]. This study calculates the difference between the original and factorized matrices, resulting in a K-component loss matrix. We then compute the mean of the absolute values of its elements to quantify information loss.

**K-component loss and Image Quality Assessment (IQA)**

According to the K-component loss, the declining trend reflects the loss variation between the NMF results and the original dataset, serving as a reference for evaluating dataset quality. As shown in Figure 2, the curve is flatter after $k = 10$, indicating that the NMF reaches its performance limit, beyond which further processing offers minimal benefit. Thus, $k = 10$ is identified as a preliminary choice for the number of components. However, to ensure this selection does not lead to overfitting, a secondary evaluation using Image Quality Assessment (IQA) is conducted.

IQA objectively analyzes and quantifies image quality through algorithms that estimate perceptual quality based on various features [34]. Its goal is to provide mathematical metrics aligned with human visual perception [35, 36]. IQA facilitates the evaluation of image quality, performance analysis of image processing algorithms, and supports decision-making for quality enhancement [37]. IQA methods are generally categorized into Full-Reference (FR) and No-Reference (NR) approaches [38]. FR-IQA, being more established, is commonly used in machine learning for image quality evaluation. It compares a reference (original) image with a target (processed or distorted) image using metrics such as Structural Similarity Index (SSIM), Peak Signal-to-Noise Ratio (PSNR), and Mean Squared Error (MSE), which are widely applied in FR-IQA [36].

**Overlapping estimation**

In 4D-STEM data extraction, the issue of overlapping arises when diffraction patterns from adjacent sample regions interfere, leading to the superposition of signals during data acquisition [24, 39]. Diffraction patterns generated at contiguous positions inherently contain contributions from neighboring areas of the sample [26]. This overlap can complicate data interpretation, particularly when analyzing subtle structural features. The primary challenge posed by overlapping in 4D-STEM lies in accurately extracting information related to the sample's local crystallography and structural properties [39]. The reliability of the reconstructed patterns heavily depends on the precision of the analytical techniques employed. Overlapping signals can introduce artifacts or inaccuracies, potentially compromising the fidelity of the final results.

4D-STEM data are first transformed into a 2D array and subsequently processed using NMF to obtain the *W* and *H* matrices. The *H* matrix ($k$, $M \times N$) represents the contribution of each basis vector from the *W* matrix to reconstructing the original matrix *V* [40]. Each row of *H* corresponds to the weights for specific data points in *V*, reflecting the extent to which each basis vector contributes to the reconstruction. In this context, *H* encapsulates how the features represented by *W* are combined to describe the original data. Here, $k$ denotes the number of clusters within the dataset, with each column indicating the weight (or probability) of a data point (diffraction pattern) belonging to a given cluster. Higher weights signify a greater likelihood of association with a specific cluster.

NMF exhibits a strong sensitivity to pertinent features during matrix factorization, effectively capturing overlapping structures within the dataset. In this context, overlaps are represented as secondary weights, while the original clusters correspond to primary weights. Utilizing the H matrix, which encapsulates all weight contributions, we extract the first and second weights and define a threshold to differentiate them. This threshold facilitates the visualization of overlapping regions within the dataset, as shown in Figure 5.

**Comparison of raw and clustered data**

Clustering analysis using NMF proves highly effective for 4D-STEM data, as it reveals latent features associated with distinct structural characteristics [25, 41]. Beyond identifying hidden patterns, NMF generates interpretable clustering results that correlate with specific crystallographic behaviors, enhancing the understanding of complex material structures [27]. Integrating clustering outputs with the original diffraction patterns not only provides validation but also validate the making-decision step for k-component determination. Indeed, aligning the original diffraction data with NMF-derived clusters strengthens the validation process, enabling a more robust assessment of cluster coherence relative to material properties and structural variations. This dual representation facilitates intuitive visualization, bridging the gap between mathematical models and structural information. Moreover, correlating NMF clusters with specific regions within the raw data allows for precise localization of structural features, thereby enriching the interpretability of clustering outcomes.

**RESULTS AND DISCUSSION**

**Impact of Dataset Filtering on NMF Robustness**

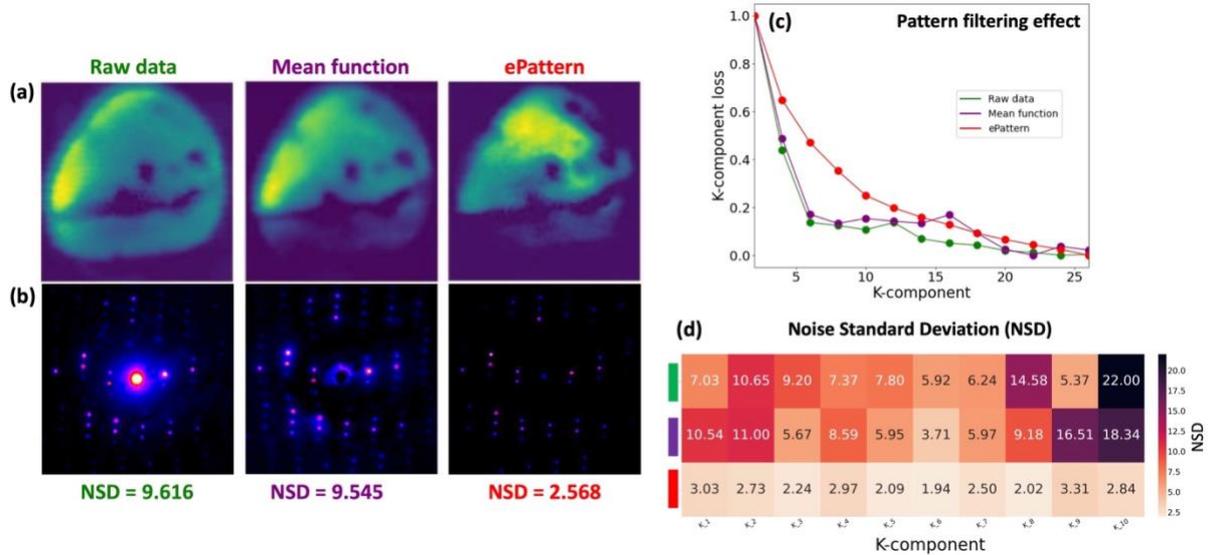

**Figure 2.** The influence of the dataset quality, with the different filtering methods, on NMF robustness. (a) The diffraction pattern and its associated mapping of the three datasets in the different quality degrees (Raw data, Mean function, and ePattern). The most representative diffraction of each dataset is executed by the NSD (noise standard deviation) method to value the consequence of filtering methods. (b) The impact of different filtering methods concerns the NMF result with the growth of components (Clustering number), the *K*-component Loss is defined as an evaluated method to measure the similarity between the dataset that needs to be dealt with NMF (V) and its corresponding NMF result (V' = W * H). (c) The NSD matrix displays the NSD value of the most typic diffraction in each clustering ($Cluster_1$ ~ $Cluster_k$) after the NMF method for the three datasets.

Applying appropriate dataset filtering techniques can significantly reduce noise, thereby enhancing the robustness of NMF [9,7]. By eliminating noisy data points, filtering ensures that NMF focuses on extracting underlying patterns rather than capturing noise, resulting in more reliable factorization outcomes. The experimental impact of dataset filtering on NMF is illustrated in Figure 2. Figure 2a demonstrates the substantial influence of filtering methods on the clustering quality of NMF results.

Three different filtering approaches are compared: (1) Raw Data: The dataset as acquired from 4D-STEM without any processing. (2) Mean Filtering: This method processes the raw data by normalizing the sum of neighboring images using a 3x3 kernel sliding across the scan [42]. This averaging technique produces a scan of unchanged size, where each Diffraction Pattern (DP) image is the average of neighboring images [42]. (3) ePattern algorithm: Proposed in our team, this novel algorithm focuses on dimensionality reduction and reconstruction of DP [42]. It employs a neural network-like structure consisting of an encoder, which extracts the most relevant features into a latent space, and a decoder, which reconstructs the diffraction patterns from the latent space representation [42]. These filtering methods highlight the importance of preprocessing in enhancing the quality and reliability of NMF results, particularly in the context of 4D-STEM data analysis.

The two proposed methods aim to enhance data quality by eliminating noisy or irrelevant data points from the dataset. Figure 2a illustrates the Noise Standard Deviation (NSD) values corresponding to each representative pattern extracted using Non-negative

Matrix Factorization (NMF). NSD serves as a metric to quantify the noise level within an image, reflecting the variation or dispersion of pixel values induced by noise [43]. Specifically, it measures the extent to which pixel values deviate from their mean due to noise interference, with higher NSD values indicating more significant noise and lower NSD values suggesting reduced noise. Consequently, by preprocessing the dataset to minimize NSD, NMF is better equipped to focus on extracting meaningful patterns from cleaner, filtered data [44].

Among the evaluated datasets, the ePattern dataset demonstrates the lowest NSD value (2.568), indicating that low-variance features have been effectively removed. Figure 2c further compares the NSD values across various K-component clusters ($\mathbf{Cluster_1} \sim \mathbf{Cluster_k}$) for different methods. The intensity of the heatmap corresponds to the magnitude of NSD, with ePattern consistently showing the lowest values (depicted by red and lightest colors). This reduction in noise enables NMF to achieve more efficient factorization and enhances the interpretability of the resulting components.

Figure 2b visualizes the impact of dataset filtering on convergence and computational efficiency during NMF processing. A comparison between raw data and preprocessed datasets (Mean function and ePattern) highlights the advantages of the latter. The ideal factorization result (V′ = W * H) is closer to the original matrix (V), with minimal deviation. Notably, the loss curve for the ePattern dataset exhibits a smooth and consistent downward trend, unlike the raw data and Mean function, which show numerous outliers. At the critical point of the steepest gradient change (k = 10), the ePattern curve demonstrates minimal fluctuation, underscoring its stability and robustness in noise handling. This improved convergence behavior facilitates more reliable and accurate NMF performance.

Moreover, by removing irrelevant features, the ePattern dataset enables NMF to produce more interpretable factors. When applied to the ePattern dataset, the resulting components represent distinct and meaningful patterns that are easier to interpret and analyze (Figure 5). In addition to superior noise reduction, the ePattern dataset enhances the stability of NMF, reduces the risk of overfitting, and prevents the model from capturing artificial patterns originating from noise [45].

**Influence of IQA on Determining the Optimal K component in NMF Analysis**

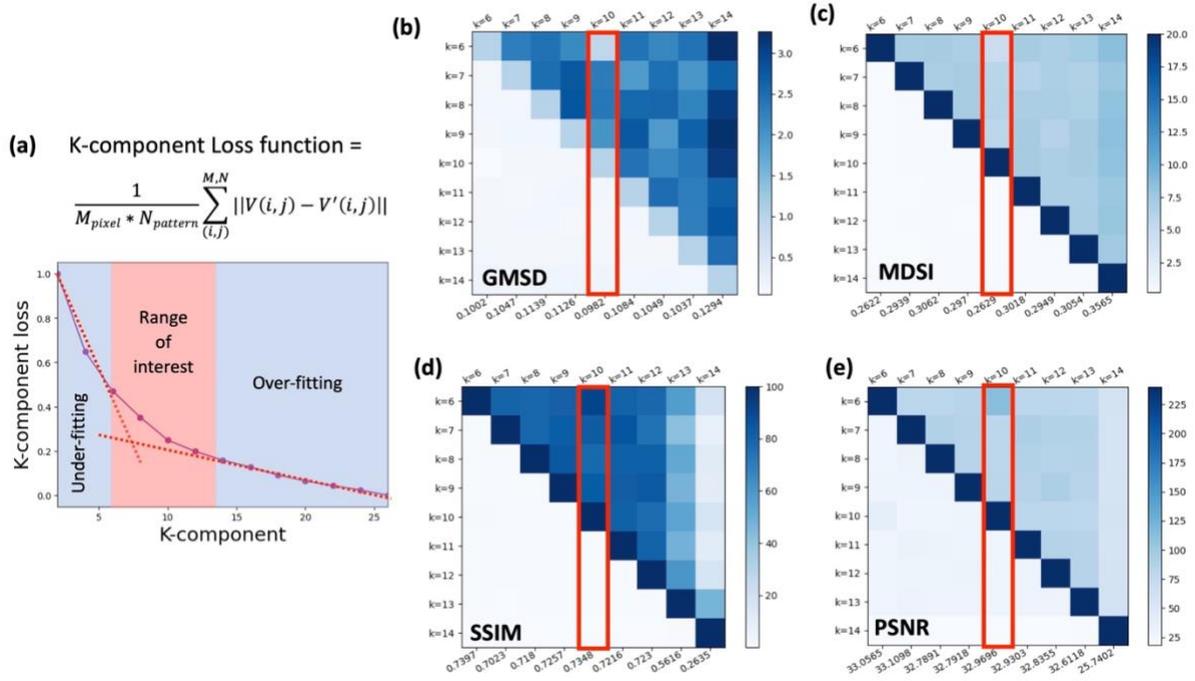

**Figure 3.** The *K*-component Loss method and its corresponding IQA (Image Quality Assessment) matrix. **(a)** The **K-component Loss method (Decision-making one)** calculates the difference between the V (NMF input matrix) and V'= W * H, each result of which depends on the choice of component (*k*). The figure demonstrates the tendency of decline according to the increase of component (*k*), the determination of the range of interest where the ideal component presumably existed in this. **(b) ~ (d),** The diffraction pattern selected by the most intense value between the k-4 and k+4 clustering is used for calculating **IQA (Image Quality Assessment)** loss for **Decision-making two**. The three matrices represent the result of the evaluation of each current diffraction pattern with the others from the Decision-making one through three different IQA algorithms in **SSIM (Structural Similarity Index Measure), GMSD (Gradient Magnitude Similarity Deviation)** and **MDSI (Mean Deviation Similarity Index)**. The objective of all metrics is to measure the similarity between two images, which aim is to determine the number of most different orientation pattern forms, namely the ideal component. **(e) PSNR (Peak Signal to Noise Ratio)** is objective to assess the degree of alteration/degradation between the diffraction pattern in the current clustering and the one in the other clustering. The result of PSNR is higher, signifying the more pertinent contribution among the totality of components.

Determining the optimal number of components k in NMF requires balancing the trade-off between reconstruction quality and model complexity [19, 21]. This involves evaluating image quality assessment (IQA) metrics and reconstruction loss across different values of k, with the objective of identifying the optimal value that provides a faithful approximation of the original data while avoiding unnecessary complexity [46]. The overarching aim is to find a value of k that effectively captures the underlying structure of the data while maintaining computational efficiency [41].

When applied to image clustering, the quality of the reconstructed images and the accuracy of the decomposition are pivotal in determining the optimal k. IQA metrics, such as Structural Similarity Index (SSIM), Peak Signal-to-Noise Ratio (PSNR), Gradient Magnitude Similarity Deviation (GMSD), and Mean Deviation Similarity Index (MDSI), are utilized to evaluate the fidelity and differences in the reconstructed images. The reconstructed data V' is expressed as V' = W * H, where each clustering operation corresponds to ($Clustering_1 = W_1 * H_1 \ldots Clustering_k = W_k * H_k$)

Figure 3a demonstrates that increasing k generally reduces reconstruction loss, as a larger number of components can theoretically capture more details of the original data. However, this also introduces the risk of overfitting, where the model begins to capture noise along with the signal. Higher values of k tend to improve IQA metrics, such as SSIM, up to a threshold, after which additional components may not enhance quality and might even degrade it due to overfitting.

The range of interest identified in Figure 3a suggests that k values between approximately 6 and 14 (centered around k = 10) achieve an optimal balance between underfitting and overfitting. Within this range, reconstruction loss decreases significantly while avoiding overfitting. This range also reflects a trade-off between capturing essential features and minimizing the incorporation of noise.

Figures 3b-3e analyze k based on four IQA algorithms. For instance, Figure 3e examines PSNR, a metric used to measure the fidelity of reconstructed images by comparing them with the original. Higher PSNR values indicate reduced distortion and noise, signifying that the NMF components have effectively captured the essential features of the original data [34]. In image compression and reconstruction contexts, PSNR values above 40 are considered excellent, whereas values below 20 are deemed unacceptable [34]. For NMF, PSNR values higher than 40 indicate that the reconstructed images retain a high degree of similarity to the original data, which is crucial for determining the optimal k.

The results suggest that k = 10 (or slightly below this value) achieves an equilibrium between preserving essential features and avoiding noise overfitting. While PSNR provides a global perspective on the fidelity of image reconstruction, other metrics like MDSI, GMSD, and SSIM complement the analysis by focusing on different aspects of image quality.

Figure 3c presents the results of MDSI, which evaluates global differences between images, including intensity and spatial information [47]. MDSI values range from -1 to 1, with higher values indicating greater similarity [47]. For NMF clustering, the goal is to maximize the distinctiveness of clusters, ensuring that diffraction patterns within clusters are noticeably distinct. At k = 8, MDSI (value = 0.3062) captures global features effectively while minimizing distortion.

In contrast, Figure 3b and Figure 3d focus on GMSD and SSIM, which measure localized and structural differences. GMSD quantifies deviations in gradient magnitudes between reference and reconstructed images, making it suitable for capturing changes in image structure caused by distortions [48, 49]. Lower GMSD values indicate greater similarity, while higher values highlight increased dissimilarity [50]. At k = 8, GMSD achieves an ideal value of 0.1139, signifying effective structural fidelity.

Concurrently, the Structural Similarity Index Measure (SSIM) provides a robust evaluation of the similarity between two images by assessing their structural information [46]. Notably, SSIM is highly sensitive to subtle structural differences, making it an effective tool for detecting slight variations between images. The SSIM index ranges from -1 to 1, where a value of 1 represents perfect structural similarity, and -1 indicates complete dissimilarity [46], [51].

In the context of clustering optimization, the analysis aims to minimize redundancy among clustering points on a global scale, with the objective of maximizing the sum of distinctly different clustering points. As illustrated in Figures 3b and 3d, based on the ultimate values for GMSD = 0.1139 and SSIM = 0.718, the analysis indicates that k = 8 represents an optimal choice for k, as further validated in Figure 4.

**Advanced Analysis and Interpretation of Orientation Mapping in 4D-STEM via NMF**

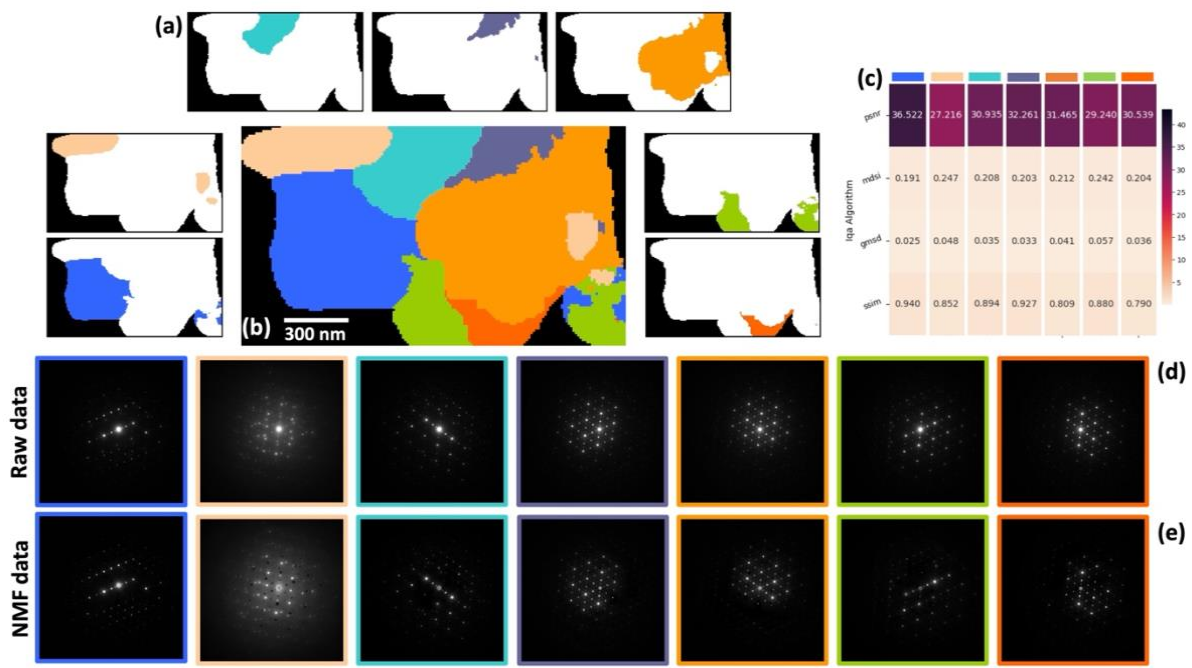

**Figure 4:** Orientation mapping results and diffraction data analysis using Non-Negative Matrix Factorization (NMF) applied to 4D-STEM datasets of LMNO cathode materials of Li-ion battery. **(a)** Segmentation maps showing distinct orientation clusters identified in the dataset. **(b)** Reconstructed orientation map over a 2 μm region, highlighting spatial variations in grain orientations. **(c)** Quantitative comparison of orientation contributions across the dataset using various NMF algorithms, including PSNR, MDSI,

GMSD, SSMI values. **(d)** Selected diffraction patterns from raw data, corresponding to identified orientation clusters. **(e)** Processed diffraction patterns after NMF decomposition, showing enhanced clarity for each orientation cluster.

In the domain of 4D-STEM, NMF serves as a powerful computational tool for decomposing diffraction pattern data into distinct structural and orientation components [27]. Here, in Figure 4, 4D-STEM analysis has been performed on LMNO cathode materials of a Li-ion battery, revealing a distinct agglomeration of crystals with noticeable overlapping between individual crystallites. This crystal configuration is inherently challenging to analyze due to the projection effects intrinsic to the TEM technique. This study emphasizes the determination of the optimal number of components (k) necessary to effectively capture and map crystallographic orientations within the dataset. By applying NMF with an optimized k = 8, the method successfully delineates and clusters distinct crystallographic orientations and phases.

The integrity of the 4D-STEM dataset, characterized by well-defined and distinct diffraction patterns, is paramount for achieving accurate component separation. Figure 4 demonstrates the robustness of NMF, validated by quantitative image quality metrics, in extracting and mapping structural features in complex materials such as cathode materials, where lattice parameters of different phases can be very close to each other. The choice of k = 8 was guided by a systematic evaluation of the trade-off between capturing essential structural details and mitigating overfitting. This optimal value was determined based on four key image quality metrics: Gradient Magnitude Similarity Deviation (GMSD), Mean Deviation Similarity Index (MDSI), Structural Similarity Index (SSIM), and Peak Signal-to-Noise Ratio (PSNR). These metrics collectively demonstrated that k = 8 achieves a balance between comprehensive feature representation and noise suppression.

The visualization in Figure 4 encapsulates the outcome of NMF applied to the dataset, where diffraction patterns from various sample regions are color-coded to represent distinct components or orientations. Each region is associated with the most representative diffraction pattern derived from the clustering process, as shown in the bottom row of the figure (labeled 1 through 8). This mapping confirms that NMF differentiates regions based on their structural similarity. The distinct colors and their corresponding diffraction patterns further validate that the selected k = 8 captures the essential crystallographic orientations and phases in the sample.

Moreover, Figure 4 underscores the critical role of dataset quality in enabling accurate component identification. High-quality diffraction patterns, characterized by sharp and well-defined features, enhance the ability of NMF to discern subtle variations in orientations with local disorientations. The sensitivity of the model to structural and orientation features at k = 8 ensures a precise balance between capturing intricate details and minimizing noise. Consequently, this approach facilitates meaningful and reliable orientation mapping, emphasizing the synergy between advanced computational techniques and high-quality experimental data.

## Identification of crystallite overlapping region

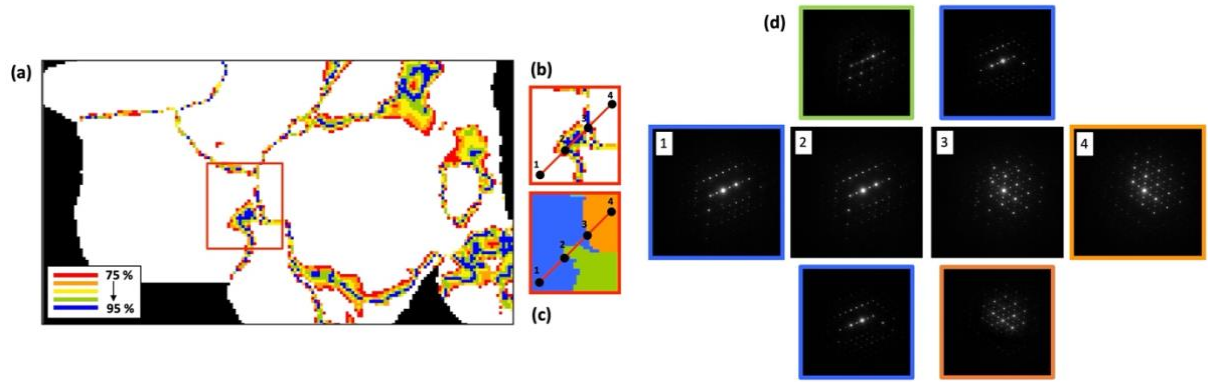

**Figure 5.** The analysis of overlapping phenomenon based upon optimal *k* (component 8). **(a)** Each line in *H* matrix represents the weights of each pixel belong to the current clustering. Each value in the weight matrix is a probability corresponding to the current clustering. **(b)** Reshaping the *H* matrix into *k*-matrix with the form (x, y), each pixel in the same position will belong to *k*-possibility, overlapping is to signify that the second weight has a significant influence in current position. **(c)** demonstrates a result of overlapping with the control of parameter thresholding **(from 75% to 95%).** Among them, the first maximum weight and the second are selected to calculate thresholding (Second Weight / First Weight) so as to detect the overlapping phenomenon.

Figure 5 demonstrates the application of NMF for the decomposition of 4D-STEM data into spatial weight matrices corresponding to individual clusters. This technique facilitates the identification and visualization of regions exhibiting significant cluster overlaps by analyzing the ratio between the second highest and maximum weights at each spatial position. Such an approach provides critical insights into the spatial distribution and interaction of clusters within the dataset.

In the context of NMF, where $V = W * H$ encodes the weight information for each cluster. Each element $H(i,j)$ represents the probability of a specific pixel belonging to a given cluster. By reshaping H into k individual weight matrices ($H_1, H_2, ..., H_k$), each matrix corresponds to a unique cluster and captures its spatial distribution as a 2D representation with dimensions (x,y).

To evaluate cluster overlap, the method systematically compares the maximum weight and the second-highest weight at each pixel location across all clusters. A ratio is computed as Second Weight/First Weight, with thresholding parameters ranging from 75% to 95% to delineate regions where the second-highest weight contributes significantly. This enables the detection of areas where clusters are not well-separated, highlighting potential overlaps.

Figure 5a illustrates the structure of H, represented as a matrix with dimensions (k ,x*y), where k is the number of clusters and x*y represents the flattened spatial dimensions of the dataset. For each spatial position (x,y), a corresponding weight vector in H indicates the likelihood of that position belonging to each cluster. For instance, if H(1,1) = 0.5, it signifies that the first diffraction pattern has a 50% likelihood of belonging to the first cluster, while H(k,1) reflects the probability of the same diffraction pattern belonging to the k-th cluster.

Following the reshaping of H into individual cluster weight matrices, Figure 5b displays these matrices (Cluster$_1$,Cluster$_2$,…,Cluster$_k$), each showing weights specific to a single cluster. Threshold values of 75%, 80%, 85%, 90%, and 95% are applied to identify regions of significant overlap. The corresponding spatial regions are then visualized using color-coded overlays to represent varying degrees of overlap.

In Figure 5c, the resulting map highlights regions of cluster overlap based on the second-weight thresholding. Different colors denote the degree of overlap, with red (75%), orange (80%), yellow (85%), and blue (95%) representing increasing thresholds. This visualization clearly delineates areas where the second-highest cluster weight plays a significant role, providing critical insights into the spatial complexity and potential interactions between clusters within the dataset.

To conclude, this section highlights the application of NMF to decompose 4D-STEM data, resolve cluster overlaps using second-to-maximum weight ratios, and visualize spatial interactions through color-coded maps.

**Comparing NMF Results with the Raw Dataset**

To assign each input dataset element to its corresponding diffraction cluster based on its index via NMF algorithm, the computed matrix H provides critical information about cluster membership. Specifically, for each column in H, $\boldsymbol{H_{(k,j)}} > \boldsymbol{H_{ij}}$ for all $i \neq k$, this indicates that the input data point Vj belongs to the k-th cluster [40]. Furthermore, the computed matrix W represents the cluster centroids, where the k-th column corresponds to the centroid of the k-th cluster [40]. Consequently, each diffraction pattern in the original dataset can be uniquely associated with a specific cluster index.

For instance, if $H_{(1,1)} > H_{(i,1)}$ (i = 2,3,4 ... *k*), this indicates that the first diffraction pattern, located at position (1,1) in the original dataset, belongs to the first cluster. Using this approach, all diffraction patterns associated with a given cluster can be identified and subsequently organized into a 3D array, where each layer corresponds to an individual diffraction image. For example, if there are N diffraction patterns of dimensions 512×512 in the first cluster, the resulting array will have dimensions 512×512×N.

To analyze these diffraction patterns further, the mean pixel intensity can be computed at each position (i, j) across all images in the cluster. This involves averaging the pixel values at position (i, j) across all NN diffraction patterns. Mathematically, the mean intensity at position (i, j) is given by:

$$Average_{(i,j)} = \frac{1}{N}\sum_{r=1}^{N} I_r(i,j)$$

Where $I_r(i, j)$ is the pixel value at position *(i, j)* in the *r*-th image.

Similarly, in the results of NMF, the element with the maximum weight in each column H(k,j) is identified. This maximum weight is then used to scale its corresponding column W(i,k). The resulting products are employed to reconstruct the diffraction pattern for the current clustering k. This process is repeated for all diffraction patterns within the current clustering, yielding a new diffraction pattern that encapsulates the characteristic information of that clustering.

Returning to the original dataset enables a comparative analysis between the initial orientations and the NMF results (see in supporting Information). This comparison not only validates the proposed method but also reinforces its effectiveness (Figure 4). Furthermore, it contributes to a deeper understanding of material characterization within the framework of clustering analysis.

**Conclusion**

In this paper, we have demonstrated a robust and systematic approach to determining the optimal number of components (k) in non-negative matrix factorization (NMF) for the analysis of 4D-STEM datasets, emphasizing the critical interplay between data quality, clustering outcomes, and computational efficiency. Through the application of various image quality assessment (IQA) metrics, including PSNR, MDSI, GMSD, and SSIM, our analysis highlights how the trade-off between reconstruction fidelity and model complexity can be effectively managed to achieve an optimal k value, with k = 8 striking the right balance between capturing essential data features and avoiding overfitting.

The integration of unsupervised multi-clustering strategies is pivotal in this context, as it facilitates a nuanced understanding of overlapping cluster structures inherent in 4D-STEM datasets. By analyzing spatial weight matrices and applying threshold-based visualization techniques, this study identified regions with significant overlap, thus enabling the identification of interaction zones and structural patterns within the data. These insights provide a more granular perspective of cluster distributions and inter-cluster relationships, which are crucial for refining decision-making processes in NMF-based analysis pipelines.

Moreover, this study underscores the importance of data preprocessing in enhancing the robustness and interpretability of unsupervised clustering results. Three preprocessing methods, Raw data, Mean function, and ePattern, were evaluated, with the ePattern method yielding the most consistent and reliable outcomes by significantly reducing noise (lower NSD values) and removing low-variance features. This demonstrates that high-quality datasets not only improve the stability of NMF results but also enable more effective multi-clustering strategies by focusing on meaningful data patterns.

Decision-making strategies in this study were further strengthened by employing IQA metrics as quantitative tools to guide the determination of k. The metrics reveal that while higher k values initially improve reconstruction accuracy, there is a threshold beyond which additional components contribute negligible quality improvements and risk overfitting. This informed decision-making approach ensures that NMF-derived results remain both computationally efficient and scientifically interpretable.

In conclusion, our study highlights a comprehensive framework that combines dataset preprocessing, unsupervised multi-clustering, and decision-making strategies to optimize NMF-based analysis of 4D-STEM datasets. By addressing overlapping cluster structures and leveraging data quality enhancements, this methodology not only improves the robustness and reliability of factorization results but also provides actionable insights into complex structural properties of cathode crystals in the 4D-STEM data. These findings establish a foundational approach for future research leveraging NMF in complex, multi-dimensional datasets and reinforce the significance of systematic preprocessing and decision-making frameworks in achieving reliable and interpretable outcomes.

## DATA AVAILABILITY

The 4D-STEM datasets used in this contribution is available for free download at https://github.com/KIRA9359/NMF

## CODE AVAILABILITY

The ePattern_Clutering is available for free download at https://github.com/KIRA9359/NMF

## ACKNOWLEDGMENTS

The research presented in this paper has received support from multiple sources. Specifically, funding has been provided by the French Research Agency (ANR) as part of the DestiNa-ion_operando project (ANR-19-CE42-0014) and by the company NanoMegas (Belgium). Additionally, the UPJV and RS2E electron microscopy platforms were utilized for this research. The authors express their gratitude to Fayçal Adrar (LRCS/RS2E) for his invaluable assistance in the testing and comparative analysis of various models in 4D-STEM data processing.

## COMPETING INTERESTS

The authors declare no competing financial or non-financial interests.

# Supporting Information

**Non-negative matrix factorization algorithm**

Lee and Seung provided the alternating optimized method to Equation (**S**2) in their seminal paper [29].

$$\min||V - WH||_F, \text{ subject to } W \geq 0, H \geq 0 \quad (2)$$

$$W^{n+1}{}_{ik} \leftarrow W_{ik} \sum_j \frac{V_{ij}}{(WH)_{ij}} H_{kj}$$

$$H^{n+1}{}_{kj} \leftarrow H_{kj} \sum_j \frac{V_{ij}}{(WH)_{ij}} W_{ik}$$

Algorithm 1 schematizes the procedure of most NMF factorization [31].

Input: Non-negative matrix ***V (M * N)*** and component ***k***

Output: ***(W, H) ≥ 0***: ***V ≈ W * H***

1. The initialization of W and H, subject (W, H) ≥ 0
2. For i = 1, 2, 3 …. max_iter do
3. $\quad W^i = update(V, H^{i-1}, W^{i-1})$
4. $\quad H^i = update(V, W^{i-1}, H^{i-1})$
5. END for

Algorithm 1: Framework of NMF Algorithm

**Figure SI_1: Dataset processing from 4D to 2D**

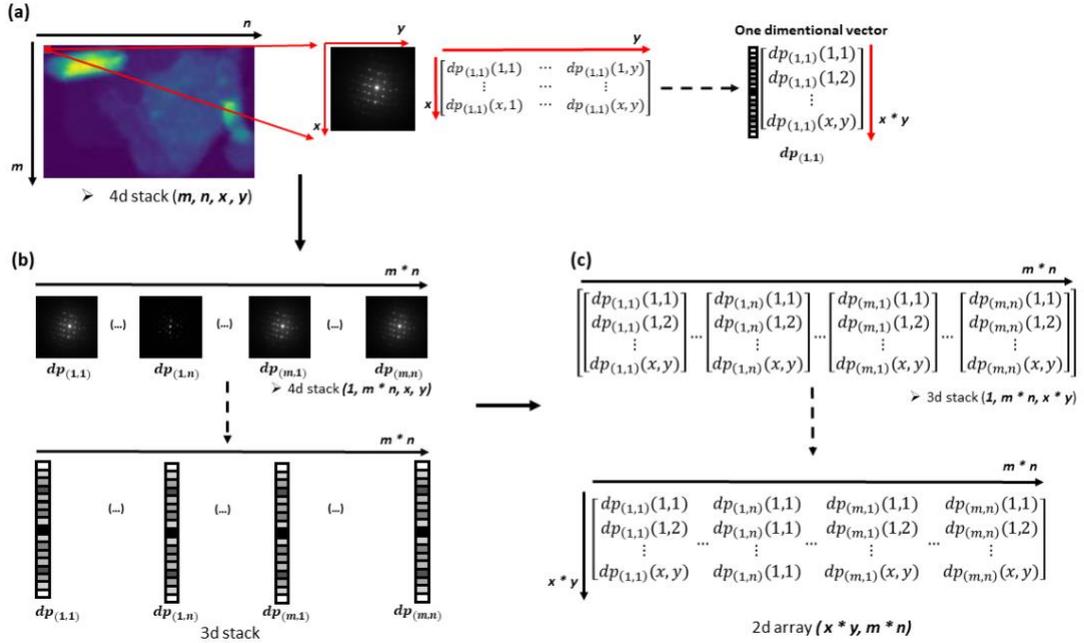

**Figure SI_1** depicts the process of transforming a 4D dataset into a 2D array, which refers to the idea from the paper 'Non-negative matrix factorization for mining big data obtained using four-dimensional scanning transmission electron microscopy' [1].

**(a)** The initial dataset is a 4D stack denoted by dimensions *(m, n, x, y)*. A 4D stack is shown on the left side with indices *m* and *n* representing the first two dimensions. Each point of *(m, n)* corresponds to a 2D diffraction pattern with dimensions *(x, y)*, as depicted in the magnified section. For example, the red point represents one pixel (one diffraction pattern) in the original dataset. $dp_{(1,1)}$ represents is the first diffraction pattern in the original dataset, moreover, $dp_{(1,1)}(1,1)$ is first pixel in the first diffraction pattern. Each 2D diffraction pattern is flattened into a 1D vector. This vector has a length of *x * y*, where each element is sequentially arranged into a single column vector, namely, from *(x, y)* to *(x * y, 1)*. **(b)** The *(m, n)* grid of 2D diffraction patterns is flattened into a single dimension, where the first dimension is 1, the second dimension is *m * n*, and the third and fourth dimensions remain as (x, y). Among them, each 2D diffraction pattern, now a 1D vector, is arranged sequentially along the new second dimension of the 3D stack. This essentially lines up the *m * n* 1D vectors side-by-side. **(c)** The 3D stack is further flattened into a 2D array. The first dimension of the 2D array corresponds to the length of the flattened 1D vectors, *x * y*. The second dimension of the 2D array corresponds to the number of such vectors, *m * n*. Finally, the 2D array is represented with dimensions (*x * y, m * n*). Specifically, each column in the 2D array represents a flattened 1D vector of a single diffraction pattern from the original 4D stack. Simultaneously, each row corresponds to a particular pixel value from the flattened diffraction patterns, the result of which is for suiting the input of NMF [1], [2].

**Algorithm SI_1: Calculation of the value of Noise Standard Deviation (NSD)**

$$\text{NSD} = \sqrt{\frac{1}{N} \sum_{i=1}^{N}(I_i - \mu)^2}$$

**Algorithm SI_1** shows the calculating of the value of **Noise Standard Deviation (NSD)**

[3] (https://ieeexplore.ieee.org/stamp/stamp.jsp?tp=&arnumber=5946565)

**NSD** is the Noise Standard Deviation. **N** is the total number of pixels in the image. $I_i$ is the pixel value at position $i$ in the image. **μ** is the average pixel value of the image. Typically, the process involves the following steps:

1. Compute the average pixel value **(μ)** of the image.
2. Calculate the squared difference between each pixel value and the average pixel value.
3. Take the average of these squared differences.
4. Finally, take the square root of this average to obtain the **NSD**.

**Algorithm SI_2: Calculation of the value of Peak Signal-to-Noise Ratio (PSNR)**

$$\text{PSNR} = 10 \, log_{10}\left(\frac{MAX^2}{MSE}\right)$$

$$\text{MSE} = \frac{1}{mn} \sum_{i=0}^{m-1} \sum_{j=0}^{n-1} [I(i,j) - R(i,j)]^2$$

**Algorithm SI_2** shows the calculating of the value of **Peak Signal-to-Noise Ratio (PSNR)** between the original image and the reference image [4]

**MAX** is the maximum possible pixel value of the image. For an 8-bit image, this is typically 255. **MSE** is the **Mean Squared Error** between the original and reference images. For the calculation of MSE, normally, $I(i,j)$ is the pixel value at position $(i,j)$ in the original image, likewise, $R(i,j)$ is the pixel value at corresponding position $(i,j)$ in the reference image. Finally, *m* and *n* are the dimensions of the images [4] [5].

**Algorithm SI_3: Calculation of the value of the Mean Deviation Similarity Index (MDSI)**

The process of calculation concerns converting the images to luminance and chromaticity channels, computing gradient and chromaticity similarities, combining these similarities, and applying deviation pooling to get the final index [6].

$$L = 0.2989R + 0.5870G + 0.1140B \quad (1)$$

$$\begin{bmatrix} H \\ M \end{bmatrix} = \begin{pmatrix} 0.30 & 0.04 & -0.35 \\ 0.34 & -0.60 & 0.17 \end{pmatrix} \begin{bmatrix} R \\ G \\ B \end{bmatrix} \quad (2)$$

**Algorithm SI_3_1** Convert the Images to Luminance and Chromaticity Channels

Algorithm SI_3_1 (1) converts the reference (R) and distorted (D) images to luminance (L) and chromaticity channels using the specified formula; Algorithm SI_3_1 (2) use the Gaussian color model to obtain the chromaticity channels $H$ and $M$ [6].

$$G_x(x) = h_x * f(x) \text{ and } G_y(x) = h_y * f(x) \quad (1)$$

$$G(x) = \sqrt{G_x^2(x) + G_y^2(x)} \quad (2)$$

$$GS(x) = \frac{2G_R(x)G_D(x) + C_1}{G_R^2(x) + G_D^2(x) + C_1} \quad (3)$$

$$F = 0.5 \times (R + D) \quad (4)$$

$$GS_{RF}(x) = \frac{2G_R(x)G_F(x) + C_2}{G_R^2(x) + G_F^2(x) + C_2} \quad (5)$$

$$GS_{DF}(x) = \frac{2G_D(x)G_F(x) + C_2}{G_D^2(x) + G_F^2(x) + C_2} \quad (6)$$

$$\hat{G}S(x) = GS(x) + [GS_{DF}(x) - GS_{RF}(x)] \quad (7)$$

**Algorithm SI_3_2** Compute Gradient Similarity (GS) and Proposed Gradient Similarity ($\hat{G}S$)

Algorithm SI_3_2 (1), (2), at first, calculate the gradient magnitudes of the luminance channels of the reference and distorted images using the Sobel operator [7]; then Algorithm SI_3_2 (3) compute the gradient similarity (GS) [6]. Algorithm SI_3_2 (4), (5), (6) will compute the fused luminance channel F and extra GS maps $GS_{RF}(x)$ and $GS_{DF}(x)$; finally, Algorithm SI_3_2 (7) calculate the proposed gradient similarity ($\hat{G}S(x)$), $C_1$, $C_2$ are constants used to control numerical stability [6].

$$\hat{C}S(x) = \frac{2H_R(x)H_D(x) + 2M_R(x)M_D(x) + C_3}{H_R^2(x) + H_D^2(x) + M_R^2(x) + M_D^2(x) + C_3}$$

**Algorithm SI_3_3** Compute Chromaticity Similarity ($\hat{C}S$)

Algorithm SI_3_3 shows that calculate the chromaticity similarity using both chromaticity channels, $C_3$ has the same objective with $C_1, C_2$ [6].

$$G\hat{C}S(x) = \alpha\hat{G}S(x) + (1 - \alpha)\hat{C}S(x)$$

**Algorithm SI_3_4** Combine Gradient and Chromaticity Similarity Maps

Algorithm SI_3_4 is to combine the gradient and chromaticity similarity maps using a weighted average, $\alpha$ is a parameter to adjust the relative importance of gradient and chromaticity similarity maps. [6].

$$MDSI = \sqrt{\frac{1}{N}\sum_{x=1}^{N}(G\hat{C}S(x) - \mu_{G\hat{C}S})^2}$$

**Algorithm SI_3_5** Apply Deviation Pooling Strategy

Algorithm SI_3_5 is to use the deviation pooling to compute the final MDSI score. Deviation pooling is based on a general formulation. Here, $\mu_{G\hat{C}S}$ is the mean of the combined similarity map values, and N is the number of pixels [6].

**Algorithm SI_4: Calculation of the value of Gradient Magnitude Similarity Deviation (GMSD)**

Gradient Magnitude Similarity Deviation (GMSD) is a perceptual image quality assessment (IQA) metric designed to measure the quality of an image by comparing its gradient magnitudes to those of a reference image [8].

$$GMS(i,j) = \frac{2G_I(i,j)G_{I'}(i,j) + C}{G_I(i,j)^2 + G_{I'}(i,j)^2 + C}$$

**Algorithm SI_4_1** Gradient Magnitude Similarity (GMS)

Algorithm SI_4_1 Compute the gradient magnitude for both the reference image and the distorted image using a gradient operator like the Sobel filter [7]. The gradient magnitude at a pixel is calculated as the square root of the sum of the squared gradients in the horizontal and vertical directions [7]. For each pixel, calculate the Gradient Magnitude Similarity (GMS) index between the reference image $I$ and the distorted image $I'$. $G_I(i, j)$ and $G_{I'}(i, j)$ are the gradient magnitudes of the reference and distorted images at pixel $(i, j)$, and C is a small constant to avoid division by zero [8].

$$GMSD = \sqrt{\frac{1}{N}\sum_{i=1}^{N}(GMS(i) - MGMS(i))^2}$$

**Algorithm SI_4_2** Gradient Magnitude Similarity Deviation (GMSD)

Algorithm SI_4_2 Mean Gradient Magnitude Similarity (MGMS) is the value of the mean of the GMS values over all pixels in the image [8]. Compute the standard deviation of the GMS values across all pixels [8]. The final GMSD value is given by the standard deviation of these GMS values [8]. N is the total number of pixels, and MGMS is the mean GMS value.

**Algorithm SI_5: Calculation of the value of structural similarity index measure (SSIM)**

The Structural Similarity Index Measure (SSIM) is a perceptual metric introduced by Zhou Wang, Alan C. Bovik, Hamid R. Sheikh, and Eero P. Simoncelli in their 2004 paper titled "Image Quality Assessment: From Error Visibility to Structural Similarity" [9]. SSIM is used for measuring the similarity between two images, aiming to quantify the visual impact of changes in image structure [9], [10].

$$SSIM(x, y) = [l(x, y)]^\alpha \times [c(x, y)]^\beta \times [s(x, y)]^\gamma$$

$$l(x, y) = \frac{2\mu_x\mu_y + C_1}{\mu_x^2 + \mu_y^2 + C_1}$$

$$c(x, y) = \frac{2\sigma_x\sigma_y + C_2}{\sigma_x^2 + \sigma_y^2 + C_2}$$

$$s(x, y) = \frac{\sigma_{xy} + C_3}{\sigma_x\sigma_y + C_3}$$

**Algorithm SI_5** The formula for SSIM between images x and y

Algorithm SI_5 SSIM combines three components: luminance ($l$), contrast ($c$), and structure ($s$):

- **Luminance**: Measures the similarity in brightness between images.
- **Contrast**: Evaluates the similarity in contrast.
- **Structure**: Assesses the similarity in structures of the images, such as edges and textures.

$\alpha, \beta, \gamma$ are parameters that define the relative importance of each component [9]. $\mu_x$ and $\mu_y$ are the means of image x and image y, $\sigma_x$ and $\sigma_y$ are the standard deviations, and $\sigma_{xy}$ is the covariance of x and y. $C_1$, $C_2$, and $C_3$ are constants to avoid instability when the value of denominators is very small [9].

**Workflow SI_5: Global plan of NMF-Clustering Analysis**

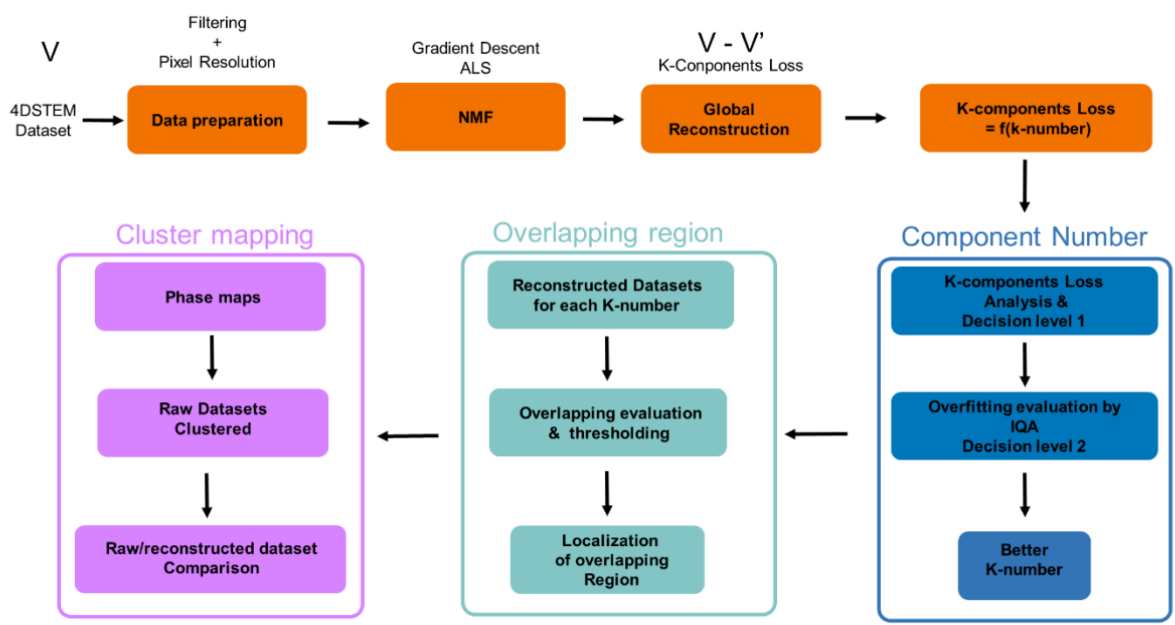

**Workflow SI_5** represents a workflow for analyzing a 4DSTEM dataset using non-negative matrix factorization (NMF). The 4DSTEM dataset $V$ through the filtering and pixel resolution adjustment to prepare the data for further analysis. The prepared data is subjected to NMF using alternating least squares (ALS) methods to decompose the dataset into two components to facilitate the later analysis. The difference between the original dataset V and the reconstructed dataset V′ is calculated through K-components loss. The K-components loss is analyzed to determine the initial decision level (Decision level 1). An overfitting evaluation using an IQA (Image Quality Assessment) method is conducted to make a refined decision on the optimal number of components (Decision level 2). This process results in the determination of a better K-number for the dataset. Once the optimal (best/ optimal) cluster number is fixed, the reconstructed datasets for different K-numbers are evaluated to identify overlapping regions. The overlapping evaluation is dependent on thresholding performed to localize the overlapping regions within the dataset. Finally, the phase maps are created from the localized overlapping regions. Simultaneously, the raw datasets are clustered based on the phase maps. A comparison is made between the raw and reconstructed datasets to ensure the accuracy and relevance of the clustering.